\documentclass{article} % For LaTeX2e
\usepackage{mlgenx2024,times}

\usepackage{amsmath,amsfonts,bm}

\def\eqref#1{equation~\ref{#1}}
\def\1{\bm{1}}

\DeclareMathAlphabet{\mathsfit}{\encodingdefault}{\sfdefault}{m}{sl}
\SetMathAlphabet{\mathsfit}{bold}{\encodingdefault}{\sfdefault}{bx}{n}

\usepackage{nicefrac}       % compact symbols for 1/2, etc.
\usepackage{microtype}      % microtypography
\usepackage{soul}           % table number reset packages end here
\usepackage{booktabs}       % professional-quality tables
\usepackage{amsfonts}       % blackboard math symbols
\usepackage{hyperref}
\usepackage{url}
\usepackage{graphicx}
\usepackage{amsmath}

\title{Integration of Graph Neural Network and Neural-ODEs for Tumor Dynamics Prediction}

\author{
    Omid Bazgir $^*$$^1$, Zichen Wang \thanks{Equal contribution} 
    \thanks{Contributed to this work during internship at Genentech.} $^1$ , Ji Won Park $^2$, Marc Hafner $^{3,4}$, James Lu $^1$ \\
    $^1$ Modeling \& Simulation/ Clinical Pharmacology, Genentech\\
    $^2$ Prescient Design/ Computational Sciences, Genentech\\
    $^3$ Oncology Bioinformatics/ Computational Sciences, Genentech\\ 
    $^4$ Discovery Oncology/ Research, Genentech\\   
    \texttt{\{bazgir.omid, wang.zichen, park.ji\textunderscore won, hafner.marc, lu.james\}@gene.com} \\
    }

\iclrfinalcopy % Uncomment for camera-ready version, but NOT for submission.
\begin{document}

\maketitle
\begin{abstract}
    In the development of anti-cancer drugs, a major scientific challenge is disentangling the complex interplay between high-dimensional genomics data derived from patient tumor samples, the organ of origin of the tumor, the drug targets associated with the specified treatments, and the ensuing treatment response. Furthermore, to realize the aspirations of precision medicine in identifying and adjusting treatments for patients depending on the therapeutic response, there is a need for building tumor dynamics models that can integrate the longitudinal tumor size measurements with multimodal, high-throughput data. In this work, we take a step towards enhancing personalized tumor dynamics predictions by proposing a heterogeneous graph encoder that utilizes a bipartite Graph Convolutional Neural networks (GCNs) combined with Neural Ordinary Differential Equations (Neural-ODEs). We apply the methodology to a large collection of patient-derived xenograft (PDX) data, spanning a wide variety of treatments (as well as their combinations) and tumor organs of origin. We first show that the methodology is able to discover a tumor dynamic model that significantly improves upon an empirical model in current use. Additionally, we show that the graph encoder is able to effectively incorporate multimodal data to enhance tumor predictions. Our findings indicate that the methodology holds significant promise and offers potential applications in pre-clinical settings.
\end{abstract}

\section{Introduction}

In the development of novel anti-cancer therapies, PDX models have become an important platform for addressing key questions, such as evaluating the treatment response to therapeutic agents and the combinations thereof, identifying the relevant biomarkers of response 
%\textcolor{blue}{identifying the relevance and consistency of biomarkers of response in \textit{in vitro} and \textit{in vivo} assays \cite{sachs2018living}}, 
and elucidating the mechanisms of resistance development \citep{byrne2017interrogating}. Furthermore, as PDX models are obtained by surgically removing patients’ tumor and implanting them in mice, co-clinical avatar trials \cite{byrne2017interrogating} can be performed, whereby treatment of the patients occur simultaneously to the treatment of the corresponding (pre-clinical) PDX models generated from the same patients. Such avatar studies facilitates real-time clinical decision making and help deliver some of the promises of precision medicine \citep{naik2023current}.

Given the myriad applications of PDX models, an important computational task is the prediction of their dynamic response to treatment from the baseline -omics data and/or early tumor size data. This entails incorporating the high-dimensional omics data measured at baseline (e.g., RNA-seq pre-treatment), with the low-dimensional but serially assessed tumor size measurements under treatment. While empirical \cite{zwep2021identification} and spline-based \cite{forrest2020generalized} tumor dynamics models  have been proposed, there has been little progress in melding such dynamic models with high dimensional omics data. \cite{zwep2021identification}, uses lasso regression to predict tumor dynamic parameters from copy-number variations (CNVs) of genes from a large PDX data set consisting of various treatments \citep{gao2015high}. In \cite{ma2021few}, a few-shot learning (multi-layer perceptron) and heterogeneous GCNs \cite{peng2022predicting}, have been proposed to learn drug responses from in-vitro (cell-line) data and predict in-vivo (PDX) outcomes within two response categories. While the predictions made by the models in \cite{ma2021few} and \cite{peng2022predicting} show promising correlations with the observed responses, they do not necessarily capture the tumor dynamics, which is crucial for clinical decision making. While their proposed methods show promise, improving the predictivity of the model via the inclusion of omics data remains an important topic for further research.

Over the past few years, Neural-Ordinary Differential Equations (NODEs) \cite{chen2018neural} has emerged as a promising deep learning (DL) methodology for making predictions from irregularly sampled temporal data. Recently, a methodology based on NODE has been developed for tumor dynamics modeling in the clinical trial setting and the generated embeddings from patients' tumor data have been demonstrated to be effective for predicting their Overall Survival (OS) \citep{laurie2023explainable}. While the Tumor Dynamic NODE (TDNODE) has set the mathematical foundations for modeling longitudinal tumor data \cite{laurie2023explainable}, a methodology for the incorporation of high-dimensional, multimodal data into such temporal models has yet to be developed.

\begin{figure}
  \centering
  \includegraphics[width=\columnwidth]{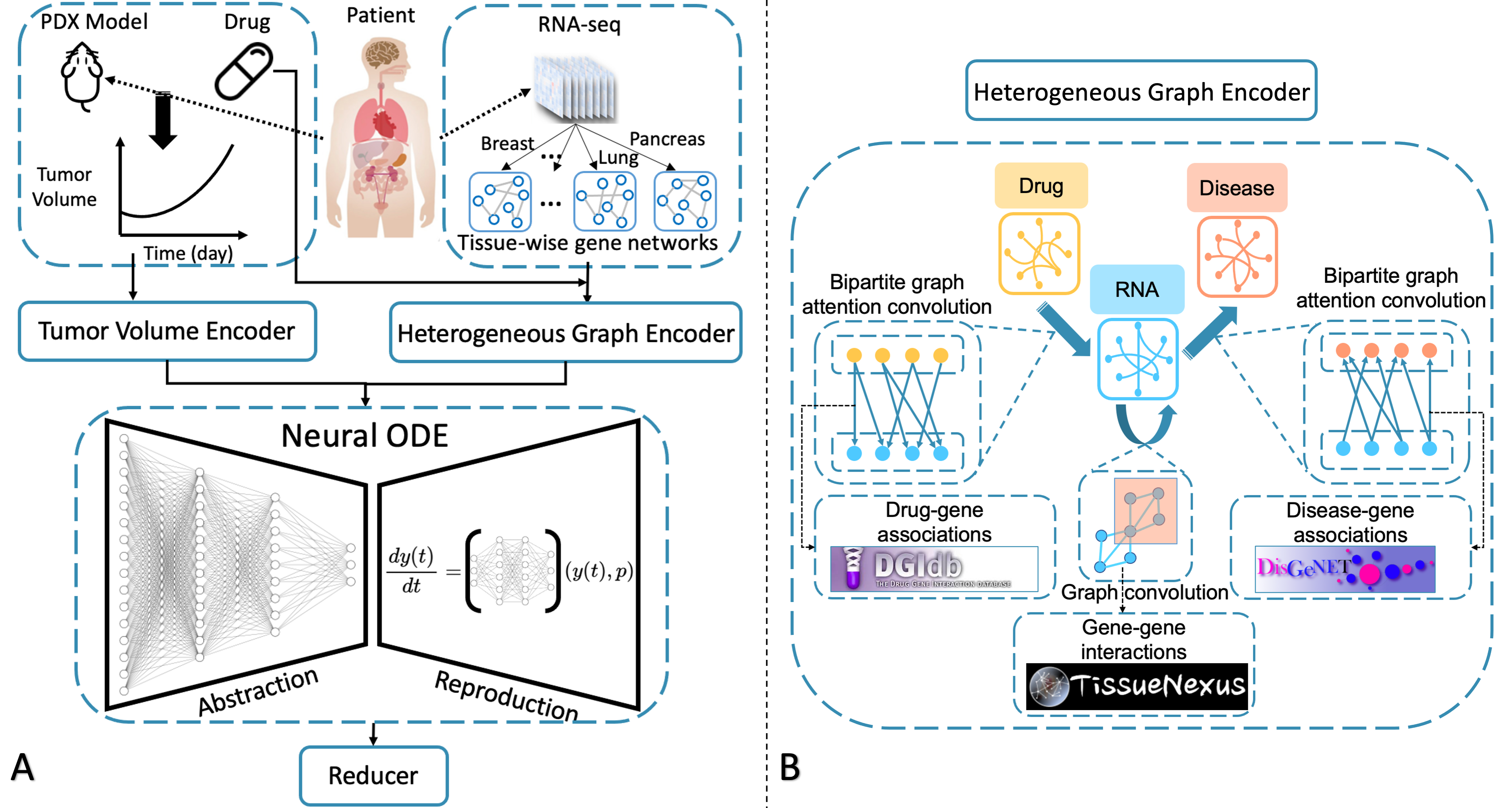} 
  \caption{\textbf{Model architecture overview.} \textbf{A)} Integration of GCNs and Neural-ODEs for tumor dynamics prediction. \textbf{B)} Heterogeneous Graph Encoder: Two bipartite graph attention convolution NNs extract disease-gene and drug-gene associations and a graph convolution NN extracts gene-gene interactions. The output of all three are integrated into the embedding representing the baseline (pre-treatment) state.} \label{SchematicFig}
\end{figure}

In this work, we propose a novel way to combine the previously developed tumor volume encoder with a heterogeneous graph encoder (see Fig.~\ref{SchematicFig}). The latter takes as inputs multimodal data consisting of drug, disease, and RNA-seq by incorporating graphs that encapsulate drug-gene associations, disease-gene associations, and gene-gene interactions. We applied our methodology to the PDX data of \cite{gao2015high} (more details provided in \autoref{pdx_data_description}). In our proposed method, we show that by leveraging the multimodal data including RNA-seq in conjunction with early tumor response data, we can significantly improve future tumor response predictions. Finally, we summarize the findings and discuss the potential future applications of the proposed approach in enabling precision medicine in oncology. 

\section{Method}
Within our multi-modal framework, we construct a multi-relational network using  three large datasets covering interactions between drugs, genes, and diseases. We use the RNA-seq data as node features within tissue-specific knowledge graphs and further integrate it with drug targets of treatments as a heterogeneous graph \citep{zhang2019heterogeneous}. 

This integration allows us to learn an embedding that captures the complex relationship among genes, tumors, and treatments. The three following graphs were utilized to represent both biological and pharmacological knowledge:
\begin{itemize}
    \item \textbf{gene-gene graph:} tissue-specific functional gene networks obtained from TissueNexus \cite{lin2022tissuenexus}, which provides gene-gene associations based on gene expression across 49 human tissues.

    \item \textbf{drug-gene graph:} obtained from DGIdb \cite{cotto2018dgidb}, which consolidates information on drug target and interacting genes from 30 disparate sources through expert curation and text mining. 

    \item \textbf{disease-gene graph:} obtained from DisGeNET \cite{pinero2016disgenet}, one of the largest collections of genes and variants associated with human diseases.
\end{itemize}

For each PDX instance in our dataset, the pre-treatment embeddings using the above multi-modal graphs are used in conjunction with the early, observed tumor measurements to predict the future tumor dynamic profiles.
The following subsections cover these two respective aspects.

\subsection{Heterogeneous Graph Encoder}

We formulate the PDX representation learning task as one of graph embedding, by fusing information from a heterogeneous network that incorporates drug, disease, and gene relationships. We apply multi-layer GCN \cite{welling2016semi} in the RNA domain to model the gene interactions (\textit{gene-gene graph}). Additionally, we use bipartite graph attention convolutions \cite{wang2020toward, nassar2018hierarchical} for message passing from drugs to target genes (\textit{drug-gene graph}), as well as gene embeddings to the disease domain (\textit{disease-gene graph}).
The mathematical formulation of our proposed framework is described as follows.

\noindent \textbf{Bipartite Graphs Attention Convolution.}
Conventional GCNs assume that all nodes belong to the same category. However, in our scenario there are heterogeneous attributes across various node types, such as genes, diseases, and drug targets. This limitation becomes evident when node attributes span across different domains. We adopt a bipartite graph as a natural representation for modeling inter-domain interactions among distinct node types. In other words, our focus is on  drug-disease interactions, rather than intra-disease or intra-drug interactions. We adapt GCNs to operate within a bipartite graph, where node feature aggregation exclusively occurs over inter-domain edges.
Specifically, let us denote graphs as $\mathcal{G} =  (\mathcal{V}, \mathcal{E})$ , where $\mathcal{V}$ represents the set of vertices, given by $\left\{v_{i}\right\}_{i=1}^{N}$, and $\mathcal{E}$ is the set of edges. We consider a bipartite graph $\mathcal{BG}(\mathcal{U}, \mathcal{V}, \mathcal{E})$  defined as a graph $\mathcal{G}(\mathcal{U} \cup \mathcal{V}, \mathcal{E})$, where $\mathcal{U}$ and $\mathcal{V}$ represent two sets of vertices (nodes) corresponding to the two respective domains. Let $u_{i}$ and $v_{j}$ denote the $i$-th and $j$-th node in $\mathcal{U}$ and $\mathcal{V}$, respectively, where $i=1,2,\dots,M$ and $j=1,2,\dots,N$. All edges within the bipartite graph exclusively connect nodes from $\mathcal{U}$ and $\mathcal{V}$ (i.e., $\mathcal{E}=\{(u, v) | u \in \mathcal{U}, v \in \mathcal{V}\}$). The features of the two sets of nodes are denoted by $X_{u}$ and $X_{v}$, where $X_{u} \in \mathbb{R}^{M \times P}$ is a feature matrix with $\vec{x}_{u_{i}} \in \mathbb{R}^{P}$ representing the feature vector of node $u_{i}$, and $X_{v} \in \mathbb{R}^{N \times Q}$ is defined similarly.
For the message passing $\operatorname{MP}_{v \rightarrow u}$ from domain $\mathcal{V}$ to $\mathcal{U}$, we define a general bipartite graph convolution ($bg$) as:
\begin{equation} \label{general BGC formula}
bg_{\mathcal{E}}(u_{i})=\rho\left(\operatorname{agg}\left(\{W_{u_{i},v_{j}}\vec{x}_{v_{j}} | v_{j} \in \mathcal{N}^{\mathcal{E}}_{u_{i}}\}\right)\right),
\end{equation}
where $\mathcal{N}^{\mathcal{E}}_{u_{i}}$ represents the neighborhood of node $u_{i}$ connected by $\mathcal{E}$ in $\mathcal{BG}(\mathcal{U}, \mathcal{V}, \mathcal{E})$ ( $\mathcal{N}^{\mathcal{E}}_{u_{i}} \subset \mathcal{V}$), $W_{u_{i},v_{j}} \in \mathbb{R}^{M \times N}$ is a feature weighting kernel transforming $N$-dimensional features to $M$-dimensional features, $\operatorname{agg}$ is a permutation-invariant aggregation operation, and the $\rho$ operator can be a non-linear activation function. In our work, we used element-wise mean-pooling and ReLU \cite{nair2010rectified} for $\operatorname{agg}$ and $\rho$ respectively.

Our bipartite graph convolution layers utilize the graph attention network \cite{velivckovic2017graph} as the backbone on the node features, resulting in the bipartite graph attention convolution layer ($bga$). Since the attention mechanism considers features of two sets of nodes, we specifically define a learnable matrix $W^{u} \in \mathbb{R}^{P \times S}$ (resp. $W^{v} \in \mathbb{R}^{Q \times S}$) for $X_{u}$ (resp. $X_{v}$), where $P,Q$, and $S$ are nodes of the message passing graphs in the respective domains. The $bga$ can be formulated as:
\begin{equation}
bga_{\mathcal{E}}(u_{i})=\operatorname{ReLU}\left(\sum_{v_{j} \in \mathcal{N}^{\mathcal{E}}_{u_{i}}}\alpha_{u_{i},v_{j}} W^{v} \vec{x}_{v_{j}}\right),
\end{equation}
The attention mechanism is a single-layer feedforward neural network, parameterized by a weight vector $\vec{a}$ and applying the LeakyReLU non-linearity function. The attention weight coefficients can be expressed as:
\begin{equation}
\alpha_{u_{i},v_{j}}=\frac{\exp\left(\rho(\vec{a}^{T}[W^{u} \vec{x}_{u_{i}} \| W^{v} \vec{x}_{v_{j}}])\right)}{\sum_{v_{k} \in \mathcal{N}^{\mathcal{E}}_{u_{i}}}\exp\left(\rho(\vec{a}^{T}[W^{u} \vec{x}_{u_{i}} \| W^{v} \vec{x}_{v_{k}}])\right)},
\end{equation}
where $T$ and $||$ represent the matrix transposition and concatenation operations respectively.

\paragraph{Information Fusion} We represent the heterogeneous network as an undirected graph $\mathcal{G}(\mathcal{V}, \mathcal{E})$ with the following three sets of nodes: drugs ($\mathcal{V}^{A}$), diseases ($\mathcal{V}^{B}$), and genes ($\mathcal{V}^{C}$). The input features of these three sets of nodes are denoted as $X_{\mathcal{V}^{A}}$, $X_{\mathcal{V}^{B}}$, and $X_{\mathcal{V}^{C}}$, respectively. The edges consist of two inter-domain sets representing drug-gene associations ($\mathcal{E}^{AC}$) and disease-gene associations ($\mathcal{E}^{BC}$), as well as an intra-domain set representing gene network connections ($\mathcal{E}^{CC}$).

First, we apply multi-layer GCNs to the RNA domain to model the gene interactions. As the \textit{gene-gene graphs} are extremely large for a graph-level prediction tasks, we first initialize the GCN model parameters by a pre-trained variational graph auto-encoder (VGAE) \citep{hu2019strategies}. The VGAE is fine-tuned on the RNA-seq data from all the PDX models in this study, where each PDX model was represented as a graph with genes as the nodes of the graph. The tumor-specific graphs were obtained from TissueNexus \citep{lin2022tissuenexus}. We used the VGAE loss function introduced in \citep{kipf2016variational}.

In the second step, we apply a single bipartite graph attention convolution layer to propagate the message from drugs to target genes. Conceptually, this step can be viewed as projecting information from the macro level (e.g., from the domain of drugs) to the micro level (e.g., the domain of genes). To formulate the message-passing step, we represent the hidden embeddings of node $v^{c}_{i}$ as $h_{v^{c}_{i}}^{(k)}$, where $k$ is the step index and when $k=0$, $h_{v^{c}_{i}}^{(0)}=x_{v^{c}_{i}}$. The term $h_{v^{c}_{i}}^{(k)}$ is computed as follows:
\begin{equation}
\operatorname{MP}_{\mathcal{V}^{A} \rightarrow \mathcal{V}^{C}}^{(k)}:h_{v^{c}_{i}}^{(k)}=bga_{\mathcal{E}^{AC}}(v^{c}_{i}) + h_{v^{c}_{i}}^{(k-1)}.
\end{equation}

Similarly, in the last step, we utilize the non-linear graph information captured by gene nodes to update the hidden embeddings of the disease nodes. In particular, we apply another bipartite graph attention convolution layer to project gene embeddings to the disease domain. Therefore, the third step can be viewed as an attentional pooling \cite{lee2019self} of the disease gene subgraph. The updated feature representations of disease nodes are computed as follows:
\begin{equation}
\operatorname{MP}_{\mathcal{V}^{C} \rightarrow \mathcal{V}^{B}}^{(k)}:h_{v^{b}_{i}}^{(k)}=bga_{\mathcal{E}^{CB}}(v^{b}_{i}) + h_{v^{b}_{i}}^{(k-1)}.
\end{equation}

We concatenate the updated drug and disease embeddings into a unified representation for PDX experiments (tumor-treatment combinations) as $\beta_{1}$, used as input features for downstream tasks.

\subsection{Tumor Volume Prediction}

We formulate a tumor volume dynamics model using a NODE which utilizes both the embedding generated from the heterogeneous graph encoder and a tumor volume encoder which takes early tumor volume data. The model aims to utilize the early tumor volume data together with baseline embedding of the baseline state (encapsulating drug, disease and RNA-seq data) in order to make personalized predictions. 

\paragraph{Tumor Volume Encoder} 
In a similar manner to \cite{laurie2023explainable}, we implement a tumor volume encoder to inform the NODE in order to make predictions tailored to a specific PDX model. This recurrent neural network (RNN) based encoder maps a short window of early observed tumor volumes (of an arbitrary length) into 
%a latent space
an embedding that we denote as $\beta_{2}$. 

\paragraph{Neural-ODE.}
In the clinical context, an approach to model tumor dynamics using NODE has been developed, demonstrating the ability of such a formalism to capture the right dynamical model from longitudinal patient tumor size data across treatment arms \citep{laurie2023explainable}. In this work, we generalize the methodology to integrate longitudinal measurements with the multimodal data in the form of an end-to-end framework. The proposed framework is demonstrated in the setting of modeling the PDX data dynamics. We consider a dynamical system of the following form:  

\begin{equation} \label{NODE_system}
    \frac{dy(t)}{dt} = f_{\theta}(y(t),\beta), t \in [0,T],
\end{equation}
where $0$ and $T$ denote the start time of PDX experimentation and the end of the prediction time respectively, $f_{\theta}$ is a neural network parameterized by a set of weights $\theta$ to be learned across all PDX data, and $\beta = [\beta_{1}||\beta_{2}]$ as the PDX embedding obtained by concatenating the outputs of the heterogeneous graph encoder and tumor volume encoder. Thus, a dynamical law represented by $f_{\theta}$ is learned across all PDX data, with the concatenated embedding $\beta$ serving to provide the initial condition for the NODE specific to the PDX instance of interest. 

After simulating \autoref{NODE_system} to obtain the time evolution of state $y(t)$, we then reconstruct the tumor volume data using a two-layer multi-layer perceptron (MLP). The NODE approach offers the benefits of accommodating variable-length sequences and varying observation intervals. It also allows us to incorporate positional encoding for capturing the temporal context. To enhance model simplicity and generalization, we keep the dimension of the ODE system ($y$ dimension) no larger than the tumor growth inhibition (TGI) model proposed by \citep{zwep2021identification}. The TGI model captured the longitudinal tumor volume measurements, per PDX, with two empirical ODEs governed by three estimated parameters: growth rate, treatment efficacy, and time-dependent resistance development. 

\section{Results}

To assess the effectiveness of the heterogeneous graph encoder in describing changes in tumor volume,
we trained the encoder using mRECIST \cite{therasse2000new,gao2015high} response labels. Specifically, we
consolidated the response categories (CR, PR, and SD) into a single response category, and PD as the second category (Table \ref{tab:mRECIST} in the Appendix). To predict these response categories, we used a binary classifier using a three-layer MLP that takes the heterogeneous graph encoder embedding as the input. The dataset was split by PDX models, and standard 5-fold cross-validation was performed. We evaluated the prediction performance using three metrics: balanced accuracy, area under the receiver operating characteristic curve (AUROC), and F1-score. Our model's performance was compared against several competing approaches, including a non-graph-based deep learning method and traditional ML methods. To investigate the impact of the pretraining strategy, we also implemented a variant of our model with a randomly initialized gene graph. More details are provided in \autoref{gcn_vs_baseline_results}. 

\begin{figure*}[htb]
\centering
\includegraphics[width=1.0\columnwidth]{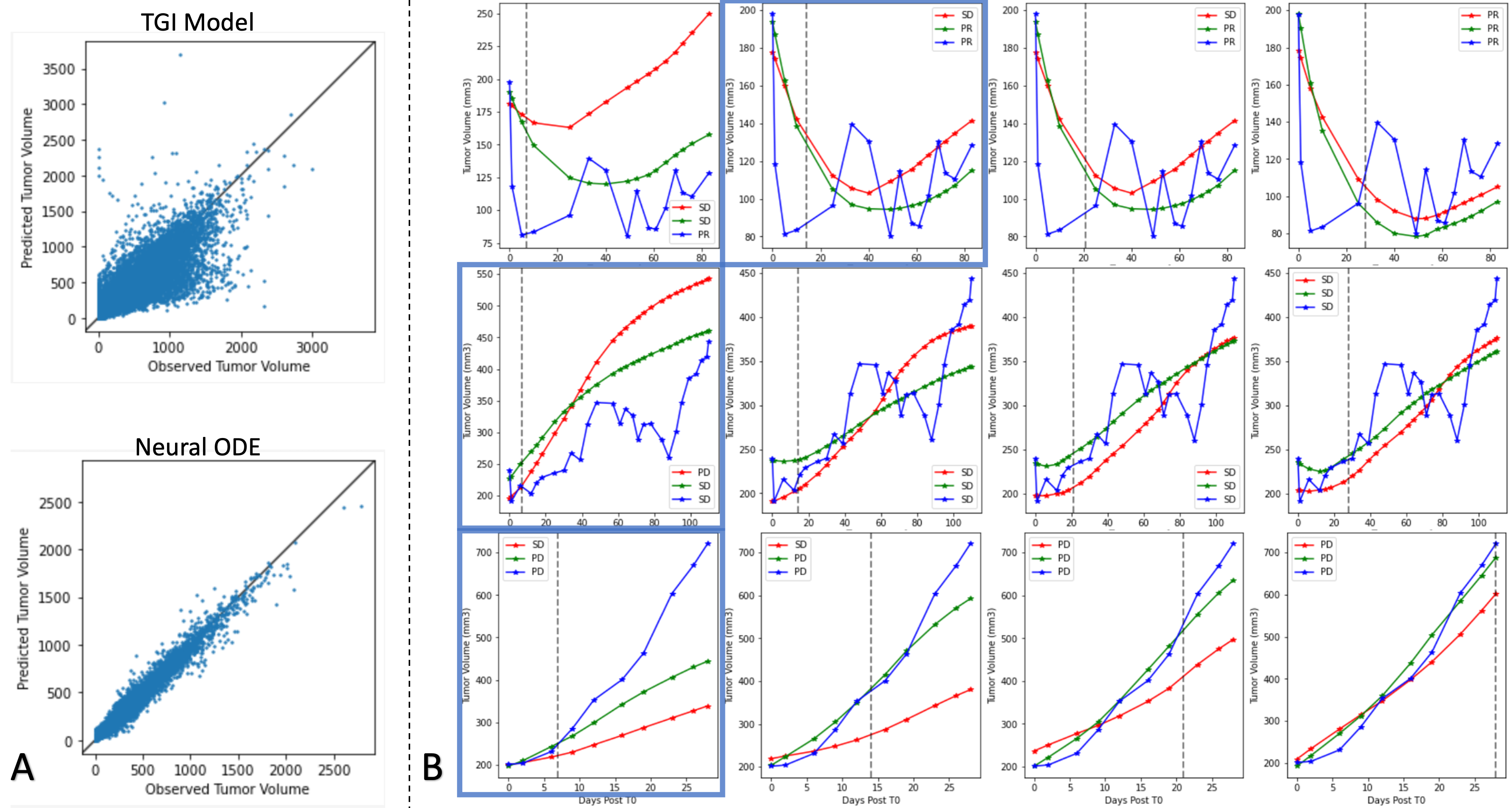}
\caption{\textbf{Tumor dynamics prediction: (A)} A comparison of tumor volume fitting between tumor growth inhibition (TGI) model proposed by \cite{zwep2021identification} and our proposed NODE model. \textbf{(B)} In each of the 3 rows: tumor volume prediction using the proposed model for an individual PDX instance is shown. The \textcolor{blue}{blue} curves represent the measurements (ground-truth), the \textcolor{green}{green} curves represent our proposed model predictions, and the \textcolor{red}{red} curves represent the TGI model predictions. Each of the column corresponds to a single choice of observation window (of increasing size from left to right): the highlighted blocks in dark blue blue indicate the observation windows beyond which our proposed model correctly captures the classification of the mRECIST response category as the ground-truth.}
\label{fig:prediction_curves}
\end{figure*}

\subsection{Reproducing Tumor Volume Data}

We evaluated the capability of our NODE in capturing the longitudinal tumor volume data, in comparison to the state-of-the-art TGI model for PDX proposed by \citep{zwep2021identification}. For this experiment, we utilized all available tumor volume data for each PDX experiment. Our approach involved encoding the longitudinal tumor volume data into a latent space using tumor volume encoder. Subsequently, we used this latent space as a part of the initial condition to solve an ODE system using the NODE model to reconstruct the dynamic tumor volume data. The population-level results are shown in Panel (A) of Figure \ref{fig:prediction_curves}. Notably, our NODE outperformed the TGI model with R2 of 0.96 compared to 0.71 and Spearman correlation of 0.96 compared to 0.86.

\subsection{Tumor Volume Dynamics Prediction}

We evaluated the ability of our model (presented in \autoref{SchematicFig}) to predict future tumor volume dynamics based on a limited observed longitudinal tumor data. We selected the observation windows of 7, 14, 21, and 28 days, to simulate real-world scenarios where early observations are used to forecast the (future unseen) tumor volume dynamics. 

We assessed the predictive performance of our model in two ways. Firstly, we employed R2 to quantify the accuracy of our model in predicting unseen tumor volumes. The results in Table \ref{Tab:r2_results} indicate the following: (1) the embedding learned from the heterogeneous graph encoder enhances the predictive performance of our proposed model; and (2) as the observation window size increases, our proposed model captures the unseen tumor dynamic more accurately. Additionally, as it is demonstrated in Panel (B) of \autoref{fig:prediction_curves}, the model effectively captures the tumor dynamic trend. 

Due to noise inherent in the tumor volume measurements and the clinical significance of mRECIST response category prediction, we also assessed our proposed model's predictive performance as a classifier. 
The mRECIST categories are derived from the predicted tumor volume time series by applying the response criteria. This evaluation measures the model's performance in correctly classifying the treatment responses based on the predicted tumor volume dynamics. \autoref{fig:classifier_results} summarizes the classification results, revealing an observable trend in which incorporating the heterogeneous graph encoder embedding improves the prediction of response categories across all observation windows. 

\section{Conclusion}
In summary, we proposed a novel approach for tumor dynamics prediction that integrates data from RNA-seq, treatment, disease as well as the longitudinal tumor volume measurements into a NODE system. In a pre-clinical PDX setting, we demonstrated that the use of  NODE vastly improves the ability of the model to capture PDX tumor data than the previously proposed TGI model, when used jointly with a graph encoder that enriches the longitudinal data. As future work, elucidating how the model predictions arise from the multimodal data using explainability techniques and/or attention weights would advance our scientific understanding of the complex interplay between gene expression profiles, tumor location, and drug targets. This methodology holds significant promise and warrants further validations, including using cancer organoids \cite{sachs2018living} and in the clinical setting.

\bibliography{mlgenx2024}
\bibliographystyle{mlgenx2024}
%%%%%%%%%%%%%%%%%%%%%%%%%%%%%%%%%%%%%%%%%%%%%%%%%%%%%%%%%%%%
\newpage
\appendix

%\section{Appendix}
{\textbf{\Large Appendix}}
\setcounter{table}{0} % Reset table counter for this section
\renewcommand{\thetable}{A.\arabic{table}} % Redefine the table numbering format
\setcounter{figure}{0} % Reset figure counter for this section
\renewcommand{\thefigure}{A.\arabic{figure}} % Redefine the figure numbering format
\setcounter{equation}{0}
\renewcommand{\theequation}{A.\arabic{equation}} % Redefine the equation
%\newline
\setcounter{section}{0}
\renewcommand{\thesection}{A.\arabic{section}} % Redefine the section
%\newline

\section{Patient-derived Xenograph (PDX) dataset}
\label{pdx_data_description}
The primary dataset utilized in this study was obtained from a large-scale pre-clinical study conducted in PDX mice models, as detailed in \cite{gao2015high}. This dataset encompassed more than 1000 PDX models, each characterized by their baseline mRNA expression levels prior to treatment. In total, the dataset covered 62 distinct treatments across six different diseases, with tumor volume measurements taken every 2-3 days. For our analysis, based on the availability of RNA-seq data we included data from 191 unique tumors and 59 different treatments, resulting in a comprehensive dataset of 3470 PDX experiments (consisting of various tumor and treatment combinations) spanning 5 tumor types. 
%This selection was based on the availability of RNA-seq data.

\section{Modified RECIST categories}
\label{mRECIST_description}

The time-dependent tumor response is determined by comparing tumor volume change at time $t$ in relation to its baseline (i.e., initial) value: 
\begin{equation}
\Delta V(t) = 100\% \times \frac{V(t) - V_{initial}}{V_{initial}}.   
\label{eq:mRECIST}
\end{equation}
The best response is defined as the minimum value of $\Delta V(t)$ for $t \geq 10$ days.

 \begin{table}[htb]
    \caption{\textbf{mRECIST categories:} The modified
    Response Evaluation Criteria in Solid Tumors (mRECIST) \cite{therasse2000new}, is a tumor progression indication being calculated in less than 64 days that "captures a combination of speed, strength and durability of response into a single value" \cite{gao2015high}. The best response is computed as the percentage of changes in the tumor volume using the Equation \ref{eq:mRECIST}. The distribution of the response categories for the dataset used in this study is provided in the third column. For further details please refer to \cite{gao2015high}}
    \label{tab:mRECIST}
    \centering
    \begin{tabular}{ccc}
        \hline
         mRECIST category & Description  & Distribution (\%) \\
         \hline
         Complete Response (CR) &  best response $\leq$ -95\% &  2.74\% \\
         Partial Response (PR)  &  -95\% $<$ best response $\leq$ -50\% & 7.12\% \\
         Stable Disease (SD) &  -50\% $<$ best response $\leq$ 35\% & 30.29\% \\
         Progressive Disease (PD) &   otherwise & 59.85\% \\
         \hline
    \end{tabular}
    
\end{table}
\section{Comparing GCN with non-GCN models.}
\label{gcn_vs_baseline} 
As the RNA-seq data is in a tabular format, therefore we use common models such as multilayer perceptron (MLP) and random forests (RF) for training and prediction on regression or classification tasks. Hence, we used MLP and RF for comparison with a complex model such as GCNs, that requires representation of the RNA-seq data as graph. Also we compared evaluated the GCN performance with and without pretraining. The reuslts are summarized in Table \ref{Tab:01}. 
\begin{table*}[htp]
    \caption{The summary of model performance on treatment response prediction using RNA-seq data. The mean and standard deviation over 5-fold cross-validation is reported.}
    \label{Tab:01}
    \centering
    \begin{tabular}{lccc}
    \hline
    Method & Balanced Accuracy & AUROC & F1 \\
    \hline
    Pretrained GCN & \textbf{\%68.1 $\pm$ 2.2} & \textbf{\%74.1 $\pm$ 2.3} & \textbf{\%62.3 $\pm$ 1.8}\\
    GCN & \%65.2 $\pm$ 1.8& \%71.3 $\pm$ 1.9& \%58.2 $\pm$ 1.8\\
    MLP & \%62.1 $\pm$ 2.3& \%68.8 $\pm$ 2.6& \%52.5 $\pm$ 2.1\\
    RF &  \%61.8 $\pm$ 1.3& \%67.6 $\pm$ 1.4& \%53.2 $\pm$ 1.3 \\
    \hline
    \end{tabular}
\end{table*}  

\begin{table}[htp]
\caption{Predictive performance of our proposed model using different observation windows quantified with R2. The mean and standard deviation over 5-fold cross-validation is reported.}
\label{Tab:r2_results}
\centering
\begin{tabular}{ccc}
\hline
Observation window & w/o graph encoder & w graph encoder \\
\hline
7 days &  \%23.3 $\pm$ 5.2& \textbf{\%30.2 $\pm$ 4.9} \\
14 days & \%45.6 $\pm$ 4.8& \textbf{\%47.9 $\pm$ 4.7} \\
21 days & \%58.6 $\pm$ 4.1& \textbf{\%60.8 $\pm$ 4.2} \\
28 days & \%65.2 $\pm$ 3.9& \textbf{\%65.9 $\pm$ 3.8} \\
\hline
\end{tabular}%
\end{table}
\newpage

\section{Results}
\subsection{GCNs outperforms baseline methods in modeling RNA-seq for treatment response prediction.}
\label{gcn_vs_baseline_results}
As shown in Table \ref{Tab:01}, GCN-based methods demonstrated superior performance in the responder classification task across all evaluation metrics. These models outperformed non-graph-based approaches by more than 9\% in F1 score and 5\% in AUROC. Additionally, we observed that our model achieved a noticeable improvement of approximately 4\% in F1 score and 3\% in AUROC when utilizing gene graph pre-training. This result highlights the effectiveness of graph reconstruction pre-training in enhancing gene latent representations. More details are provided in \autoref{gcn_vs_baseline}.

\subsection{mRECIST categories prediction with and without heterogeneous graph encoder}
\begin{figure*}[htb]
\centering
\includegraphics[width=1.0\columnwidth]{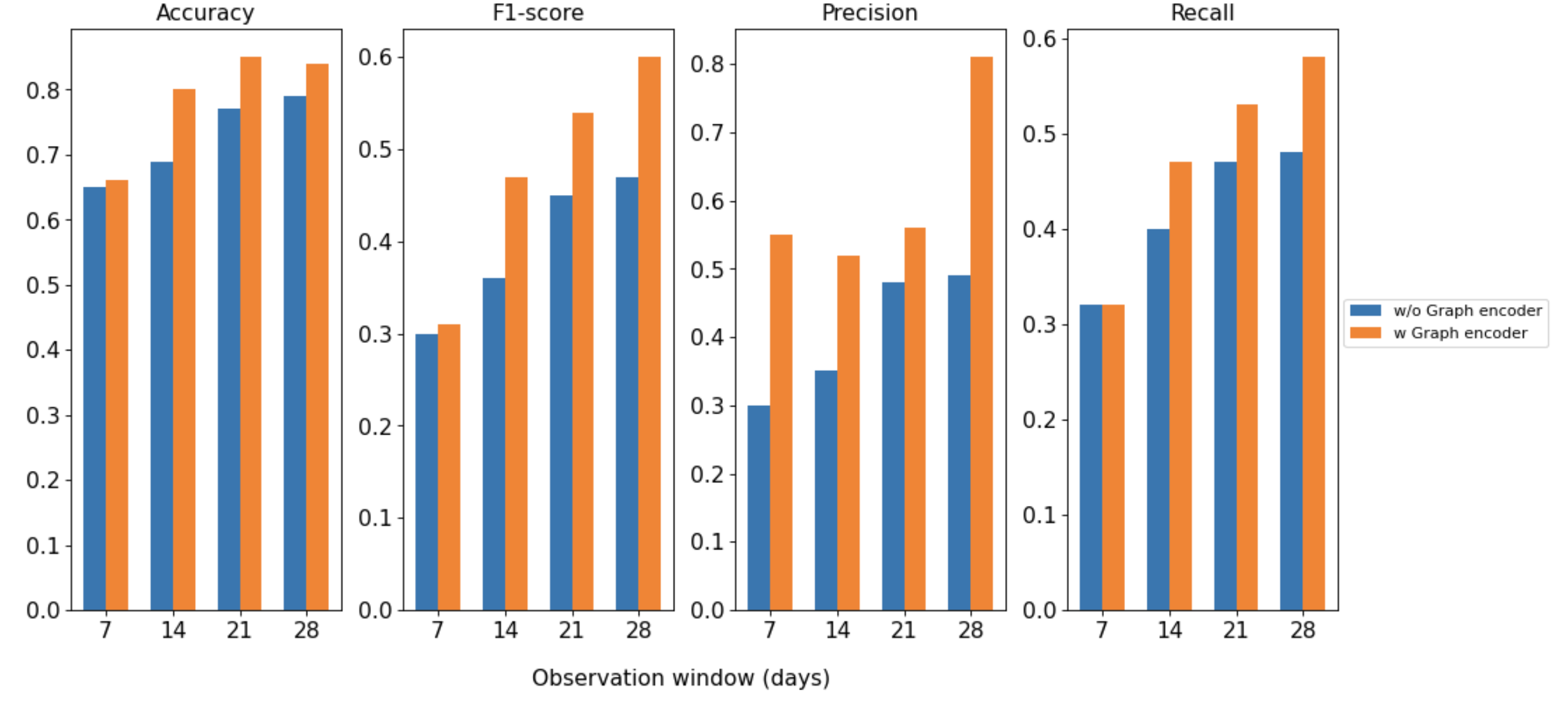}
\caption{Predictive performance of our proposed model as a classifier for mRECIST categories, with and without the heterogeneous graph encoder and considering different lengths of observation windows.} 
\label{fig:classifier_results}
\end{figure*}

\section{Training procedures}
\subsection{Pretrained GCN}
We used VGAE with two GCN layers to pretrain the \textit{gene-gene graph} using RNA-seq data and knowledge graph. We used 3 different losses for the pretraining procedure; node reconstruction, edge reconstruction, and KL-divergence. The node reconstruction loss is the mean squared error (MSE) of genes (nodes of the input graph) and reconstructed genes (decoder output). The edge reconstruction loss is the cross entropy of positive and negative edges in the graph. The KL-divergence is used as a regularization term in the loss function to assure continuity and completeness in the latent space.

\subsection{Treatment response prediction using RNA-seq dataset}
In this experiment we defined the problem as a classification task and used cross entropy loss function for the pretrained GCN, GCN, and MLP. For the RF model we used \textit{Gini} criteria as a measure of impurity.

\subsection{Treatment response prediction using Heterogeneous Graph}
The treatment response prediction experiment using the heterogeneous graph neural networks was conducted as a regression task to predict the entire tumor volume for each PDX model. Then the predicted tumor volumes were converted into response categories using the Table \ref{mRECIST_description}. This end-to-end model takes all 3 modalities including RNA-seq, drug, and treatment data to the heterogeneous graph encoder, and the observed tumor volume data was fed into the tumor volume encoder. Then the predicted tumor volume from the Neural-ODE is used to be compared with the ground truth tumor volume data of the entire tumor volume using MSE loss function.

\end{document}